\documentclass[runningheads]{llncs}
\usepackage[T1]{fontenc}
\usepackage{graphicx}
\usepackage{tikz}
\usepackage{algorithm}
\usepackage{algpseudocode}
\usepackage{amsmath,amssymb}
\usepackage{booktabs}
\usepackage{bm}
\usepackage{epsfig}
\usepackage{subfig}
\usepackage{times}
\usepackage{url}

\newcommand{\ie}{i.e.}
\newcommand{\eg}{e.g.}
\newcommand{\etal}{et al. }

\graphicspath{{images/}}

\begin{document}

\title{Semi-Supervised Variational Adversarial Active Learning via Learning to Rank and Agreement-Based Pseudo Labeling}

\titlerunning{Semi-Supervised Variational Adversarial Active Learning}
%
\author{Zongyao Lyu\orcidID{0000-0002-7542-5818} \and
William J. Beksi\orcidID{0000-0001-5377-2627}}
%
\institute{The University of Texas at Arlington, Arlington TX 76019, USA}

\maketitle              

\begin{abstract}
Active learning aims to alleviate the amount of labor involved in data labeling
by automating the selection of unlabeled samples via an acquisition function.
For example, variational adversarial active learning (VAAL) leverages an
adversarial network to discriminate unlabeled samples from labeled ones using
latent space information. However, VAAL has the following shortcomings: (i) it
does not exploit target task information, and (ii) unlabeled data is only used
for sample selection rather than model training. To address these limitations,
we introduce novel techniques that significantly improve the use of abundant
unlabeled data during training and take into account the task information.
Concretely, we propose an improved pseudo-labeling algorithm that leverages
information from all unlabeled data in a semi-supervised manner, thus allowing
a model to explore a richer data space. In addition, we develop a ranking-based
loss prediction module that converts predicted relative ranking information
into a differentiable ranking loss. This loss can be embedded as a rank
variable into the latent space of a variational autoencoder and then trained
with a discriminator in an adversarial fashion for sample selection. We
demonstrate the superior performance of our approach over the state of the art
on various image classification and segmentation benchmark datasets.

\end{abstract}

\section{Introduction}
\label{sec:introduction}
Deep learning has shown impressive results on computer vision tasks mainly due
to annotated large-scale datasets. Yet, acquiring labeled data can be extremely
costly or even infeasible. To overcome this issue, active learning (AL) was
introduced \cite{cohn1996active,settles2009active}. In AL, a model is
initialized with a relatively small set of labeled training samples. Then, an
AL algorithm progressively chooses samples for annotation that yield high
classification performance while minimizing labeling costs. By demonstrating a
reduced requirement for training instances, AL has been applied to various
computer vision applications including image categorization, image
segmentation, text classification, and more.

\renewcommand{\floatpagefraction}{.9}
\begin{figure}
\centering
\includegraphics[width=0.8\textwidth]{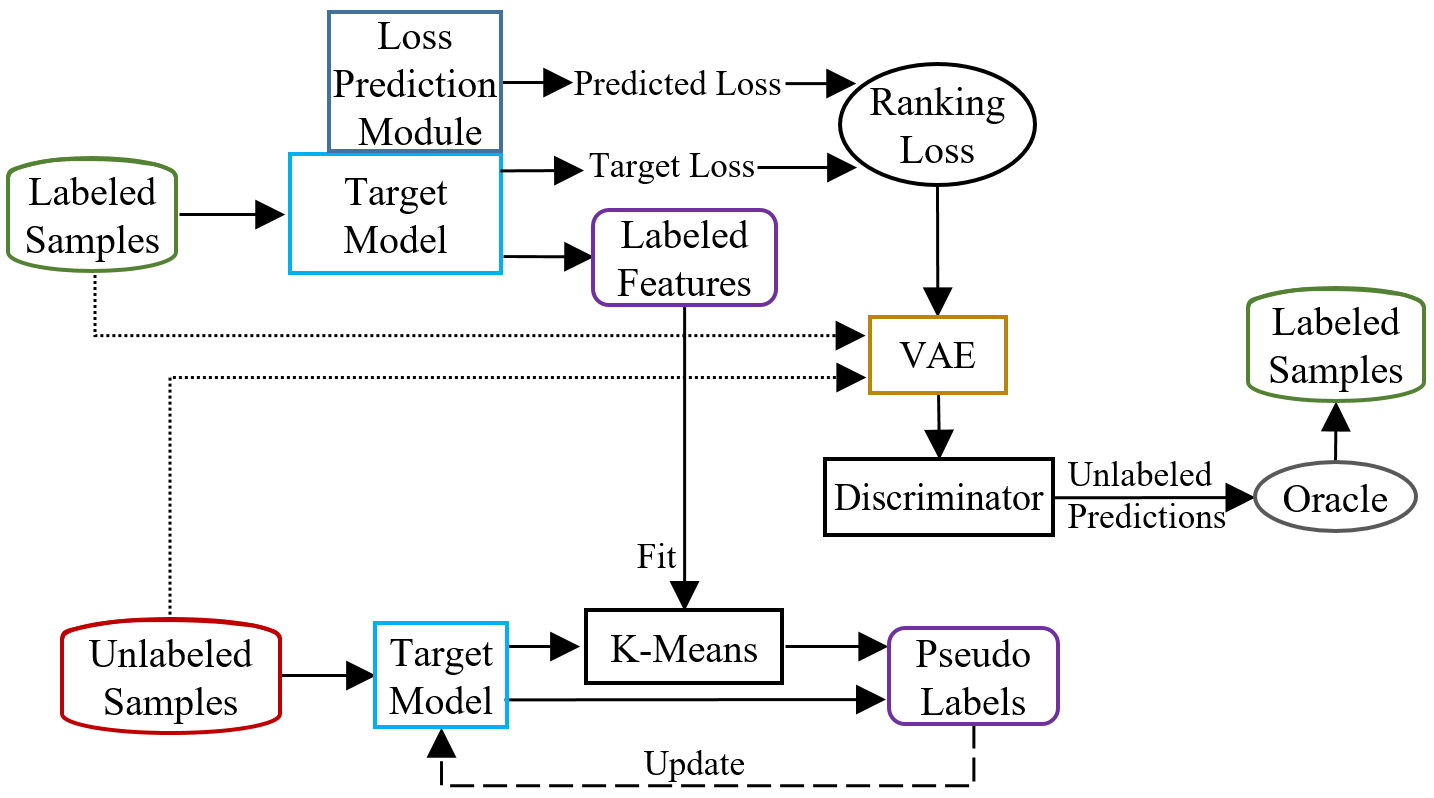}
\caption{An overview of SS-VAAL. First, a loss prediction module attached to
the target model predicts losses on the input data. Next, the predicted losses
along with the actual target losses are transformed into ranking losses via a
pretrained ranking function. Unlabeled samples are then passed to the target
model and subsequently through a k-means algorithm to acquire pseudo labels for
additional training. Finally, a discriminator following a variational
autoencoder is trained in an adversarial manner to select unlabeled samples for
annotation.}
\label{fig:overview}
\end{figure}

Among the most prevalent AL strategies, pool-based approaches have access to a
huge supply of unlabeled data. This provides valuable information about the
underlying structure of the whole data distribution, especially for small
labeling budgets. Nevertheless, many AL methods still fail to leverage valuable
information within the unlabeled data during training. On the other hand,
semi-supervised learning (SSL), in particular the technique of pseudo labeling,
thrives on utilizing unlabeled data. Pseudo labeling is based on the concept
whereby a model assigns ``pseudo labels'' to samples that produce
high-confidence scores. It then integrates these samples into the training
process. In contrast, AL typically selects only a handful of highly-informative
samples (\ie, samples with low prediction confidence) at each learning step and
regularly seeks user input. Although AL and pseudo labeling both aim to
leverage a model's uncertainty, they look at different ends of the same
spectrum. Hence, their combination can be expected to achieve increased
performance \cite{gilhuber2022verips}.


\textit{In light of this observation, we propose to exploit both labeled and
unlabeled data during model training by (i) predicting pseudo labels for
unlabeled samples, and (ii) incorporating these samples and their pseudo labels
into the labeled training data in every AL cycle.} The idea of using unlabeled
data for training is not new. Earlier work by Wang \etal \cite{wang2016cost}
showed promising results by applying entropy-based pseudo labeling to AL.
However, pseudo labeling can perform poorly in its original formulation. The
subpar performance is attributed to inaccurate high-confidence predictions made
by poorly calibrated models. These predictions produce numerous incorrect
pseudo labels \cite{arazo2020pseudo}. To tackle this issue, we introduce a
novel agreement-based clustering technique that assists in determining pseudo
labels. Clustering algorithms can analyze enormous amounts of unlabeled data in
an unsupervised way \cite{nguyen2004active,coletta2019combining}, and cluster
centers are highly useful for querying labels from an oracle
\cite{huang2021uncertainty}.
Our two-step process involves (i) separately clustering labeled and unlabeled
data, (ii) assigning each piece of unlabeled data an initial pseudo label and a
clustering label.
A final pseudo label is confirmed only if these two labels agree. The end result
is a significant reduction in the number of incorrect pseudo labels.

The second aspect of our work focuses on the sample selection strategy in AL.
We base our approach on the VAAL \cite{sinha2019variational} framework. VAAL
uses an adversarial discriminator to discern between labeled and unlabeled data,
which informs the sample selection process. Later adaptations of VAAL (\eg,
TA-VAAL \cite{kim2021task}) incorporate a loss prediction module, relaxing the
task of exact loss prediction to loss ranking prediction.  Additionally, a
ranking conditional generative adversarial network (RankCGAN)
\cite{saquil2018ranking} is employed to combine normalized ranking loss
information into VAAL. To better integrate task-related information into the
training process, we propose a learning-to-rank method for VAAL. This decision
is inspired by the realization that the loss prediction can be interpreted as a
ranking problem \cite{li2021learning}, a concept central to information
retrieval. We refine the loss prediction process by applying a contemporary
learning-to-rank technique for approximating non-differentiable operations in
ranking-based scores. The loss prediction
module estimates a loss for labeled input, 
converting the predicted loss and actual target loss into a differentiable
ranking loss. This ranking loss, along with labeled and unlabeled data, is
provided as input into an adversarial learning process that identifies unlabeled
samples for annotation. 
Therefore, by explicitly exploiting the loss information directly related to the
given task, task-related information is integrated into the AL process. The
architecture of our proposed method, SS-VAAL, is depicted in
Fig.~\ref{fig:overview}.

To summarize, our contributions are the following.
\begin{enumerate}
  \item We create a novel agreement-based pseudo-labeling technique that
  optimally harnesses rich information from abundant unlabeled data in each AL
  cycle, capitalizing on the advantages of unsupervised feature learning.
  \item We devise an enhanced loss prediction module that employs a
  learning-to-rank method, yielding a more effective sample selection strategy.
  We develop a ranking method that explicitly ranks the predicted losses by
  taking into account the entire list of loss structures, as opposed to only the
  pairwise information considered in prior approaches.
  \item We highlight the superior efficacy of our approach through its
  application to common image classification and segmentation benchmarks.
\end{enumerate}
Our source code is publicly available \cite{ss-vaal}.

\section{Related Work}
\label{sec:related_work}
\subsection{Active Learning}
AL methods operate on an iterative principle of constructing a training set.
This involves (i) cyclically training the classifier on the current labeled
training set, and (ii) once the model converges, soliciting an oracle (\eg,
human annotator) to label new points selected from a pool of unlabeled data
based on the utilized heuristic. This type of AL belongs to pool-based AL, in
which our methodology lies. Pool-based AL can be classified into three groups:
(i) uncertainty (informativeness-based) methods
\cite{lewis1995sequential,gal2017deep,beluch2018power}, (ii) diversity
(representativeness-based) methods \cite{sener2018active}, and (iii) hybrid
methods \cite{huang2014active,yang2015multi,ash2020deep,yan2022clustering}
based on the instance selection strategy they use. Among the various instance
selection strategies, uncertainty-based selection is the most prevalent. It
measures the uncertainties of new unlabeled samples using the predictions made
by prior classifiers.


Diversity-based AL methods rely on selecting a few examples by increasing the
diversity of a given batch. The core-set technique \cite{sener2018active} was
proposed to minimize the distance between the labeled and unlabeled data pool
using the intermediate feature information of a convolutional deep neural
network (DNN) model. It was shown to be an effective method for large-scale
image classification tasks and was theoretically proven to work best when the
number of classes is small. However, as the number of classes grows the
performance deteriorates.

AL methods that combine uncertainty and diversity use a two-step process to
select high-uncertainty points as the most informative points in a batch. Li
\etal \cite{li2013adaptive} presented an adaptive AL approach that combines an
information density and uncertainty measure together to label critical
instances for image classification. Sinha \etal \cite{sinha2019variational}
observed that the uncertainty-based batch query strategy often results in a
lack of sample diversity and is vulnerable to outliers. As a remedy, they
proposed VAAL, a method that utilizes an adversarial learning approach to
distinguish the spatial coding features of labeled and unlabeled data, thereby
mitigating outlier interference. It also employs both labeled and unlabeled
data to jointly train a variational autoencoder (VAE) in a semi-supervised
fashion.
Sample selection in VAAL is based on the prediction from the discriminator
adversarially trained with the VAE. While VAAL incorporates unlabeled data
during the adversarial learning process, it neglects this data during target
task learning due to its inherently task-agnostic nature. An extended version
of VAAL \cite{zhang2020state} was proposed to combine task-aware and
task-agnostic approaches with an uncertainty indicator and a unified
representation for both labeled and unlabeled data.

Task-aware VAAL (TA-VAAL) \cite{kim2021task} is an alternative extension of
task-agnostic VAAL that combines task-aware and task-agnostic approaches.
TA-VAAL adapts VAAL to consider the data distribution of both labeled and
unlabeled pools by combining them with a learning loss approach
\cite{yoo2019learning}. The learning loss is a task-agnostic method. It
includes a loss prediction module that learns to predict the target loss of
unlabeled data and selects data with the highest predicted loss for labeling.
TA-VAAL relaxes the task of learning loss prediction to ranking loss prediction
and employs RankCGAN to incorporate normalized ranking loss information into
VAAL. However, the main difference between VAAL and TA-VAAL is the use of
task-related information for learning the ranking function in conjunction with
information from unlabeled data. Even so, unlabeled data is not directly
applied to target task learning. \textit{To rectify this, we propose a novel
pseudo-labeling technique that can be integrated into each AL cycle, enabling
the comprehensive utilization of the rich information contained within
unlabeled data for direct learning of the target task.} Another recent method,
multi-classifier adversarial optimization for active learning (MAOAL)
\cite{geng2023multi}, employs multiple classifiers trained adversarially to
more precisely define inter-class decision boundaries while aligning feature
distributions between labeled and unlabeled data. We demonstrate that our
method outperforms MAOAL in image classification tasks.

\subsection{Semi-Supervised Learning}
SSL is a strategy that leverages both labeled and unlabeled data for model
training, with an emphasis on utilizing abundantly available unlabeled data.
Several techniques have been proposed to exploit the relationship between
labeled and unlabeled data to achieve better performance. A notable technique
is pseudo labeling \cite{lee2013pseudo} where a model, once trained, is used
to predict labels for unlabeled data. These pseudo-labeled data are then used
in subsequent training iterations. Other methods, such as multi-view training
\cite{sindhwani2005beyond} and consistency regularization
\cite{sajjadi2016regularization}, leverage the structure or inherent properties
of the data to derive meaningful information from the unlabeled portion.

Several efforts have been made to combine SSL and AL methods to make better use
of the unlabeled data during training
\cite{wang2016cost,simeoni2020rethinking,buchert2022exploiting}. A common
strategy in this integrated approach is to apply pseudo labeling techniques
during each AL cycle. This enriches the training set and improves model
accuracy by combining SSL's efficient use of unlabeled data with AL's selective
querying, offering a cost-effective solution for scenarios with limited labeled
data. Although simple to implement, pseudo labeling can perform relatively
poorly in its original formulation. The underperformance of pseudo labeling is
generally attributed to incorrect high-confidence predictions from models that
are not properly calibrated. This causes a proliferation of wrong pseudo
labels, thus resulting in a noisy training process \cite{rizve2021in}.
\textit{Our enhanced pseudo-labeling approach addresses this limitation by
incorporating unsupervised feature learning through the use of clustering.}
Clustering algorithms are employed to group the unlabeled data, and the cluster
centers are used for verifying the predicted pseudo labels. This greatly
reduces the number of incorrect pseudo labels as the labels are assigned based
on the proximity to cluster centers, which represents the classes better than
individual instances.



\section{Method}
\label{sec:method}
Let $(X_L, Y_L)$ be a pool of data and their labels, and $X_U$ the pool of
unlabeled data. Training starts with $K$ available labeled sample pairs
$(X_L^K, Y_L^K)$. Given a fixed labeling budget in each AL cycle, $b$ samples
from the unlabeled pool are queried according to an acquisition function. Next,
the samples are annotated by human experts and added to the labeled pool. The
model is then iteratively trained on the updated labeled pool $(X_L^{K+b},
Y_L^{K+b})$, and this process is repeated until the labeling budget is
exhausted.

SS-VAAL enhances the VAAL framework and its variant, TA-VAAL, as follows. VAAL
employs adversarial learning to distinguish features of labeled and unlabeled
data, which reduces outlier impact and leverages both labeled and unlabeled
data in a semi-supervised training scheme. TA-VAAL, building on the groundwork
of VAAL, utilizes global data structures and local task-related information for
sample queries.
Our methodology improves upon these predecessors by harnessing the full
potential of the data distribution and model uncertainty, hence further
refining the query strategy in the AL process.

\begin{figure}
\centering
\includegraphics[width=12cm]{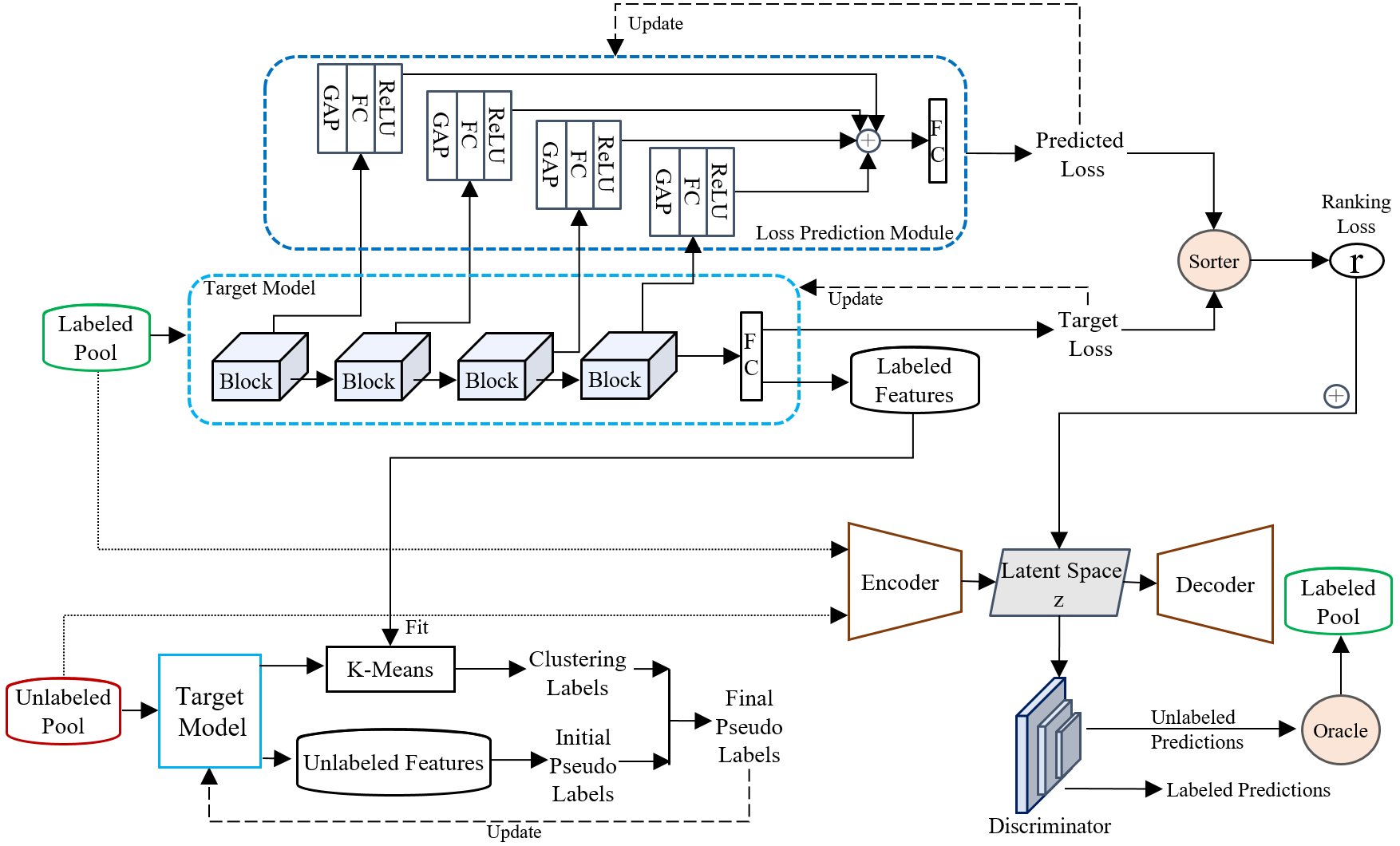}
\caption{The detailed architecture of SS-VAAL. (Stage 1) A loss prediction
module is attached to the target model to predict losses on the input data.
These predicted losses, along with the actual losses obtained from the target
model, are transformed into ranking losses via a pretrained ranking function.
Features of the labeled samples are extracted from the target model to fit a
k-means algorithm. (Stage 2) Unlabeled samples are processed through the target
model to obtain initial pseudo labels. The k-means algorithm, already fit with
labeled features, is also applied to the unlabeled samples to obtain clustering
labels for them. Initial pseudo and clustering labels are combined to determine
the final pseudo labels. These unlabeled samples and their pseudo labels are
then used for additional training of the target model. (Stage 3) Both labeled
and unlabeled samples are fed into an encoder network to learn the latent
variables. The learned and rank variables are trained adversarially with a
discriminator. Sample selection is based on the predicted probability from the
discriminator.}
\label{fig:architecture}
\end{figure}

\subsection{Clustering-Assisted Pseudo Labeling}
Both VAAL and TA-VAAL do not fully use unlabeled data in the target learning
task. Therefore, we propose to exploit both types of data during model training
as follows.
Given $X_L$ and $X_U$ for labeled and unlabeled examples, respectively, we
apply a classifier $f$ on the unlabeled data $f(X_U)$, and select and assign
pseudo labels $\hat{y}$ for the most certain predictions. Traditionally, the
labeled set will be directly augmented by $y = y + \hat{y}$ for the next round
of
training.
Nonetheless, pseudo labeling in its initial form may produce high-confidence
predictions that are incorrect, resulting in numerous erroneous pseudo labels
and ultimately causing an unstable training process.

\textit{To mitigate this issue, we present a semi-supervised pre-clustering
technique for each pseudo label selection process that enhances robustness by
reducing incorrect pseudo labels.} In each AL cycle, we first train a model on
the available labeled data. We modify the network to output both the
probability score and the feature vector from the last fully-connected layer
before sending it to the softmax function. Then, we fit a k-means clustering
algorithm on the output features of the labeled training data. This allows the
algorithm to learn the structure of the labeled data and predict clusters each
of whose centroid corresponds to one of the classes of the dataset. One thing
to note is that the cluster assignments won't necessarily correspond directly
to the classes of the dataset being trained. This is because clustering
algorithms (\eg, k-means) do not have any inherent knowledge of class labels
and thus the cluster labels they assign have no intrinsic meaning. To be
meaningful, we map the clustering labels to the actual classes to ensure that
they correspond to each other. This is done by assigning each cluster label to
the most frequent true class label within that cluster based on the labeled
training data.

Next, we train a classifier on all unlabeled data to get the predicted
probability vectors
\begin{equation}
  \sum^U \mathbf{p}(y_i = j \,|\, \mathbf{x}_i) = f(X_U) \xrightarrow{} \mathcal{R}^{c},
\end{equation}
where $c$ is the number of total classes. We assign initial pseudo labels to
the unlabeled data with the most certain predictions only when their associated
probabilities are larger than a threshold $\tau$ (we set $\tau$ = 0.95 in the
experiments), i.e.,
\begin{gather}
  j^{*} = \max\limits_{j} \mathbf{p}(y_i = j \,|\, \mathbf{x}_i), \notag \\
  \hat{y}_i = \begin{cases}
  \arg j, &j > \tau \\
  0, &otherwise.
  \end{cases}
\label{eq:pseudo_labeling}
\end{gather}
Then, we apply the k-means function learned on the labeled data to the
unlabeled data to predict the clusters they belong to. Each unlabeled sample is
grouped to the nearest cluster and assigned a label to which the cluster
centroid corresponds.

Each unlabeled data point will now have both an initial pseudo label and a
clustering label. Lastly, we compare the temporary pseudo labels with the
clustering labels to determine a final pseudo label for each unlabeled data
only if they agree with each other. By doing so, we reduce the number of
incorrect pseudo labels, thus taking full advantage of the abundant unlabeled
data for model training. Stage 2 in Fig.~\ref{fig:architecture} shows this
agreement-based pseudo-labeling process. We demonstrate improvement over
conventional pseudo labeling through an ablation study in the supplementary
material.

\subsection{Loss Prediction with Learning-to-Rank}
In LL4AL \cite{yoo2019learning}, Yoo and Kweon designed a loss prediction
module attached to the target network and jointly learned to predict the losses
of unlabeled inputs. The loss is predicted as a measure of uncertainty,
directly guiding the sample selection process.
LL4AL has proven to be effective, yet the ``loss-prediction loss'' that is key
to this approach is not trivial to calculate. The loss module adapts roughly to
the scale changes of the loss instead of fitting to the exact value. Similar to
TA-VAAL, we incorporate task-related information into the learning process by
combining VAAL with the loss prediction module. \textit{Unlike TA-VAAL, which
employs a GAN-based ranking method to address this issue, our approach
integrates VAAL with a listwise learning-to-rank technique that explicitly
ranks the predicted losses thus taking into account the entire list of loss
structures.} This decision stems from the observation that learning the loss
prediction can be seen as a ranking problem. Additionally, the loss in TA-VAAL
resembles the original LL4AL as both only consider the neighboring data pairs
and ignore the overall list structure. This motivates us to use a more
appropriate listwise ranking scheme.
Ranking is crucial for many computing tasks, such as information retrieval, and
it is often addressed via a listwise approach (\eg,
\cite{cao2007learning,liu2009learning}). This involves taking ranked lists of
objects as instances and training a ranking function through the minimization
of a listwise loss function defined on the predicted and ground-truth lists
\cite{xia2008listwise}.

\begin{figure*}[h]
\centering
\includegraphics[width=0.65\textwidth]{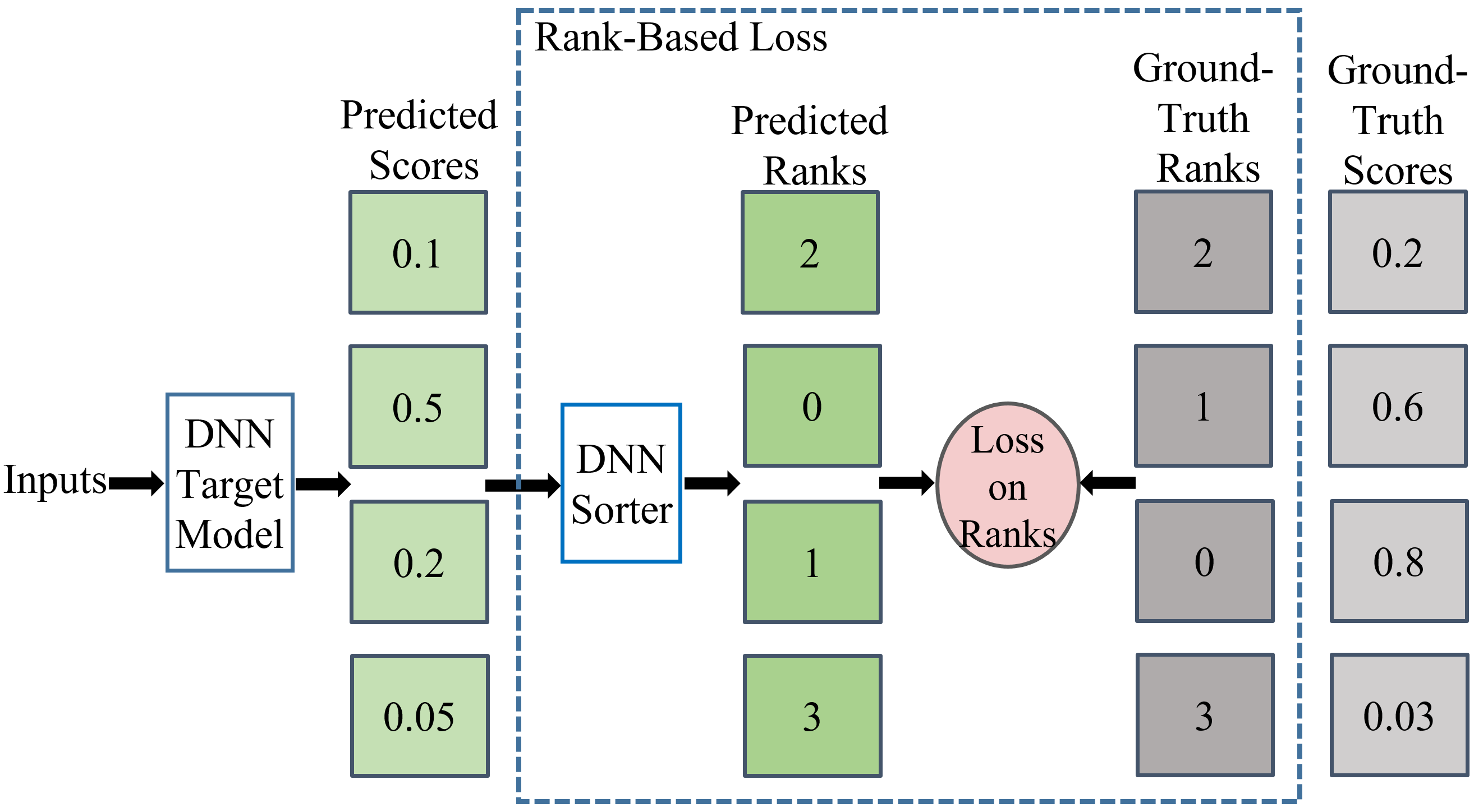}
\caption{An overview of the SoDeep sorter architecture. A pretrained
differentiable DNN sorter converts the raw scores into ranks given by the
target model. A loss is then applied to the predicted rank, backpropagated
through the differentiable sorter, and used to update the weights.}
\label{fig:sodeep}
\end{figure*}

SoDeep \cite{engilberge2019sodeep} is a method for approximating the
non-differentiable sorting operation in ranking-based losses. It uses a DNN as
a sorter to approximate the ranking function and it is pretrained separately on
synthetic values and their ground-truth ranks. The trained sorter can then be
applied directly in downstream tasks by combining it with an existing model
(\eg, the loss prediction module) and converting the value list given by the
model into a ranking list. The ranking loss between the predicted and
ground-truth ranks can then be calculated and backpropagated through the
differentiable sorter and used to update the weights of the model.
Fig.~\ref{fig:sodeep} illustrates the sorter architecture. We find this process
works well with the loss prediction task in the loss module. \textit{Therefore,
we apply SoDeep to the loss prediction module and learn to predict the ranking
loss as a variable that injects task-related information into the subsequent
adversarial learning process, which increases the robustness of the unlabeled
sample selection.} Concretely, we substitute the loss prediction module into
the sorter architecture as the DNN target model to produce the predicted scores
where the target losses are used as the ground-truth scores.

The upper-right side of Fig.~\ref{fig:architecture} displays the architecture
of the modified loss learning process. We retain the basic structure of the
original loss prediction module. Given an input, the target model generates a
prediction, while the loss prediction module takes multi-layer features as
inputs that are extracted from multiple mid-level blocks of the target model.
These features are connected to multiple identical blocks each of which
consists of a global average pooling layer and a fully-connected layer. Then,
the outputs are concatenated and passed through another fully-connected layer
to be converted to a scalar value as the predicted loss $\mathcal{L}_{pred}$.
The target prediction and annotation are used to calculate a loss
$\mathcal{L}_{target}$, which assists in training the target model. This target
loss is treated as the ground-truth loss for the loss prediction module and
used to compute the loss-prediction loss. Specifically, the predicted loss and
the target loss are passed through the pretrained SoDeep sorter and converted
to a differentiable ranking loss
\begin{equation}
  \mathcal{L}_{ranking} = SoDeep(\mathcal{L}_{pred}, \mathcal{L}_{target}),
\end{equation}
which can be used to update the weights of the model. The objective function of
the task learner with the ranking loss module is
\begin{equation}
  \mathcal{L} = \mathcal{L}_{target}(\hat{y}_L, y_L) + \lambda \mathcal{L}_{ranking},
\end{equation}
where $\hat{y}_L$ and $y_L$ are the predicted and ground-truth labels,
respectively, and $\lambda$ is a scaling constant. This training process is
illustrated as Stage 1 in Fig.~\ref{fig:architecture}. The learned ranking loss
is embedded as a task-related rank variable in the latent space of a VAE for
the subsequent adversarial learning process, which is described in detail in
Sec.~\ref{subsec:joint_training_with_a_variational_autoencoder_and_discriminator}.
Stage 1 of the two-stage training is summarized in
Alg.~\ref{alg:model_training}.

\begin{algorithm}
\begin{algorithmic}[1]
\Require{Labeled data pool $(X_L, Y_L)$, unlabeled data pool $X_U$, pretrained
SoDeep sorter $S$, initialized model $\theta_T$, training epochs $N$, threshold
$\tau$}
\Ensure{}
\For{$i = 1$ to $N$}
\State{Train target model $\theta_T$ on labeled data $(X_L, Y_L)$ to obtain features
and target loss $\mathcal{L}_{target}$}
\State{Obtain predicted loss $\mathcal{L}_{pred}$ through loss prediction module
by fusing multi-level features}
\State{$\mathcal{L}_{ranking} \gets S(\mathcal{L}_{target}, \mathcal{L}_{pred})$}
\State{Fit features to k-means algorithm}
\State{Apply k-means on unlabeled data $X_U$ and predict clustering labels (CL)}
\State{Predict initial pseudo labels (IPL) $\hat{y}_i$ for unlabeled data $X_U$
using \eqref{eq:pseudo_labeling}}
\State{Final pseudo labels $\gets$ IPL $\cap$ CL}
\State{Train model on labeled and pseudo labeled data}
\EndFor
\State{\Return{Trained model $\theta_T$}}
\end{algorithmic}
\caption{Target Model Training}
\label{alg:model_training}
\end{algorithm}

\subsection{Joint Training with a Variational Autoencoder and Discriminator}
\label{subsec:joint_training_with_a_variational_autoencoder_and_discriminator}
For sample selection, we extend VAAL by utilizing a VAE and an adversarial
network (discriminator) to distinguish labeled from unlabeled data.
\textit{Unlike VAAL, which only considers the data distribution for adversarial
learning, we incorporate task-related information by embedding the ranking loss
as a rank variable in the latent space for training both the VAE and the
discriminator.}
Let $p_\theta$ and $q_\phi$ be the encoder and decoder parameterized by
$\theta$ and $\phi$, $\mathbf{z}_L$ and $\mathbf{z}_U$ the latent variables
generated from the encoder for labeled and unlabeled data, and $r_L$ the rank
variable for the labeled data. Let $p(\mathbf{z}) = \mathcal{N}(\mathbf{0}, I)$
be the unit Gaussian prior. The transductive learning of the VAE to capture
latent representation information on both labeled and unlabeled data is
characterized by
\begin{equation}
\begin{split}
  \mathcal{L}_{VAE}^{trans} = \mathbb{E}[\log q_\phi(\mathbf{x}_L\,|\,\mathbf{z}_L, r_L)]
  - \beta KL(p_\theta(\mathbf{z}_L\,|\,\mathbf{x}_L)||p(\mathbf{z})) \\
  +\, \mathbb{E}[\log q_\phi(\mathbf{x}_U\,|\,\mathbf{z}_U, \hat{l}_U)]
  - \beta KL(p_\theta(\mathbf{z}_U\,|\,\mathbf{x}_U)||p(\mathbf{z})),
\end{split}
\label{eq:trans_vae}
\end{equation}
where $\hat{l}_U$ is the predicted loss $\mathcal{L}_{pred}$ over unlabeled
data, $\beta$ is the Lagrangian parameter, and $\mathbb{E}$ denotes the
expectation \cite{higgins2017beta}.

With the latent representations $\mathbf{z}_L$ and $\mathbf{z}_U$ learned by
the VAE of both the labeled and unlabeled data, the objective function of the
VAE in adversarial training is then
\begin{equation}
  \mathcal{L}_{VAE}^{adv} = - \mathbb{E}[\log(D(p_\theta(\mathbf{z}_L\,|\,\mathbf{x}_L, r_L)))] - \mathbb{E}[\log(D(p_\theta(\mathbf{z}_U\,|\,\mathbf{x}_U, \hat{l}_U)))].
\label{eq:adv_vae}
\end{equation}
Combining \eqref{eq:trans_vae} and \eqref{eq:adv_vae}, the overall objective
function of the VAE is
\begin{equation}
  \mathcal{L}_{VAE} = \mathcal{L}_{VAE}^{trans} + \eta \mathcal{L}_{VAE}^{adv},
\label{eq:vae}
\end{equation}
where $\eta$ is a coefficient hyperparameter. The objective function of the
discriminator $D$ during adversarial training is
\begin{equation}
  \mathcal{L}_{D}^{adv} = - \mathbb{E}[\log(D(p_\theta(\mathbf{z}_L\,|\,\mathbf{x}_L, r_L)))] - \mathbb{E}[\log(1 - D(p_\theta(\mathbf{z}_U\,|\,\mathbf{x}_U, \hat{l}_U)))],
\label{eq:adv_disc}
\end{equation}
and the overall objective function of the adversarial training is
\begin{equation}
  \min_{p_\theta}\max_{D} \mathbb{E}[\log(D(p_\theta(\mathbf{z}_L\,|\,\mathbf{x}_L, r_L)))] + \mathbb{E}[\log(1 - D(p_\theta(\mathbf{z}_U\,|\,\mathbf{x}_U, \hat{l}_U)))].
\label{eq:adv}
\end{equation}

\begin{algorithm}
\begin{algorithmic}[1]
\Require{Labeled data $(X_L, Y_L)$, unlabeled data $X_U$, rank
variable (\ie, ranking loss) $r_L$, trained model $\theta_T$, initialized models
$\theta_{VAE}$ and $\theta_D$, training epochs $N$, labeling budget $b$}
\Ensure{}
\For{$i = 1$ to $N$}
\State Compute $\mathcal{L}_{VAE}^{trans}$, $\mathcal{L}_{VAE}^{adv}$, and
$\mathcal{L}_{VAE}$ using \eqref{eq:trans_vae}, \eqref{eq:adv_vae}, and
\eqref{eq:vae}, respectively
\State Compute $\mathcal{L}_{D}^{adv}$ using \eqref{eq:adv_disc}
\State Update $\theta_{VAE}$ and $\theta_D$ using \eqref{eq:adv}
\State Select samples $X_b$ with $\min_b{D(X_U)}$
\State Query labels for $X_b$: $Y_b \gets Oracle(X_b)$
\State $(X_L, Y_L) \gets (X_L, Y_L) \cup (X_b, Y_b)$
\State $X_U \gets X_U - X_b$
\EndFor
\State \Return{Updated $(X_L, Y_L), X_U$}
\end{algorithmic}
\caption{Adversarial Training and Sample Selection}
\label{alg:adversarial_training}
\end{algorithm}

The VAE and discriminator are trained in an adversarial manner. Specifically,
the VAE maps the labeled $p_\theta(\mathbf{z}_L\,|\,\mathbf{x}_L)$ and
unlabeled $p_\theta(\mathbf{z}_U\,|\,\mathbf{x}_U)$ data into the latent space
with binary labels 1 and 0, respectively, and tries to trick the discriminator
into classifying all the inputs as labeled. On the other hand, the
discriminator tries to distinguish the unlabeled data from the labeled data by
predicting the probability of each sample being from the labeled pool. Thus,
the adversarial network is trained to serve as the sampling scheme via the
discriminator by predicting the samples associated with the latent
representations of $z_L$ and $z_U$ to be from the labeled pool $x_L$ or the
unlabeled pool $x_U$ according to its predicted probability $D(\cdot)$. In
short, sample selection is based on the predicted probability of the
discriminator adversarially trained with the VAE. The smaller the probability,
the more likely the sample will be selected for annotating. This adversarial
training process is shown as Stage 3 in Fig.~\ref{fig:architecture} and
summarized in Alg.~\ref{alg:adversarial_training}.

\section{Experiments}
\label{sec:experiments}
To evaluate the proposed SS-VAAL framework, we carried out extensive
experiments on two computer vision tasks: image classification and semantic
segmentation.

\subsection{Active Learning for Image Classification}
\textbf{Datasets.} To evaluate SS-VAAL, we performed experiments on the
following commonly used datasets: CIFAR-10, CIFAR-100
\cite{krizhevsky2009learning}, Caltech-101 \cite{fei2004learning}, and ImageNet
\cite{deng2009imagenet}. Both the CIFAR-10 and CIFAR-100 datasets consist of
50,000 training images and 10,000 test images that are $32 \times 32$ in size.
The Caltech 101 dataset contains 9,146 images, split between 101 different
object categories. Each object category contains between 40 and 800 images,
each of which is approximately $300 \times 200$ pixels. ImageNet is a
large-scale dataset with more than 1.2 million images from 1,000 classes.

\noindent \textbf{Implementation details.} We first trained a SoDeep sorter to
rank the losses. Given the close performance of several available sorter
options, we opted for the LSTM sorter. The sorter was trained with a sequence
length of 128 for 300 epochs on synthetic data consisting of vectors of
generated scalars associated with their ground-truth rank vectors. This
training is separate from the AL process. After training was complete, the
sorter was applied to the loss prediction module to convert the predicted and
target losses into ranking losses for the AL process.

For CIFAR-10 and CIFAR-100, we applied the same data augmentation as the
compared methods, including a $32 \times 32$ random crop from $36 \times 36$
zero-padded images, random horizontal flip, and normalization with the mean and
standard deviation of the training set. The target model underwent 200 epochs
of training on labeled data with a batch size of 128, then 100 epochs of
semi-supervised training on pseudo-labeled data. The initial learning rate was
set to 0.1, and reduced to 0.01 and 0.001 at 160 and 240 epochs, respectively.
For training, we employed ResNet-18 \cite{he2016deep} as the target network
with the loss prediction module described in Sec.~\ref{sec:method} using
stochastic gradient descent with the momentum set to 0.9 and a weight decay of
0.0005. Experiments began with an initial labeled pool of 1000 / 2000 images
from the CIFAR-10 / CIFAR-100 training set, respectively. At each stage, the
budget size was 1000 (CIFAR-10) / 2000 (CIFAR-100) samples. The pool of
unlabeled data consisted of the residual training set from which samples were
selected for labeling by an oracle. Upon labeling, these samples were
incorporated back into the initial training set and the process was carried out
again on the updated training set.

For Caltech-101 and ImageNet, the images were resized to $224 \times 224$ and
we initiated the process with 10\% of the samples from the dataset as labeled
data with a budget size equivalent to 5\% of the dataset. All other settings
remained the same as those used for CIFAR-10 and CIFAR-100, except that the
main task was trained for 100 epochs for the ImageNet dataset. The
effectiveness of our approach was assessed based on the accuracy of the test
data. We compared against a random sampling strategy baseline and
state-of-the-art methods including the core-set approach
\cite{sener2018active}, LL4AL \cite{yoo2019learning}, VAAL
\cite{sinha2019variational}, TA-VAAL \cite{kim2021task}, and MAOAL
\cite{geng2023multi}.

\noindent \textbf{Results.} All the compared against methods were averaged
across 5 trials on the CIFAR-10, CIFAR-100, and Caltech-101 datasets, and
across 2 trials on ImageNet. 
Fig.~\ref{fig:classification} and Fig.~6 
(see supplementary material) show the classification accuracy on the benchmark
datasets. The results obtained for the competing methods are largely in line
with those reported in the literature. Our comprehensive methodology, SS-VAAL,
incorporates both the ranking loss prediction module and the
clustering-assisted pseudo labeling. The empirical results consistently show
that SS-VAAL surpasses all the competing methods at each AL stage.

\begin{figure}[h]
\centering
\setlength{\abovecaptionskip}{0.11cm}
\subfloat[]{
  \includegraphics[width=.35\columnwidth]{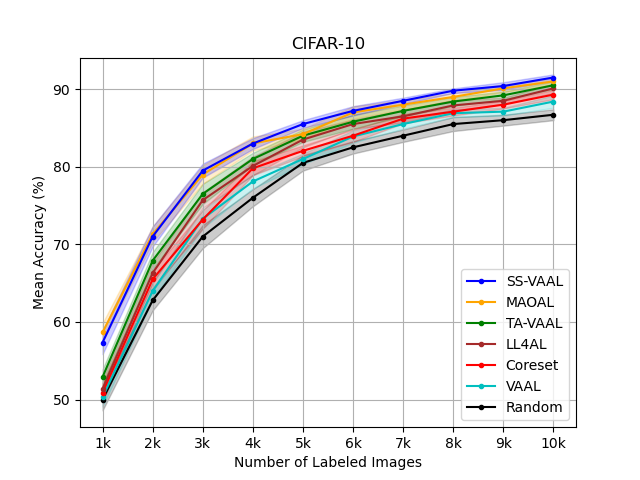}\hspace{-2mm}
  \label{subfig:CIFAR10_results}}
\subfloat[]{
  \includegraphics[width=.35\columnwidth]{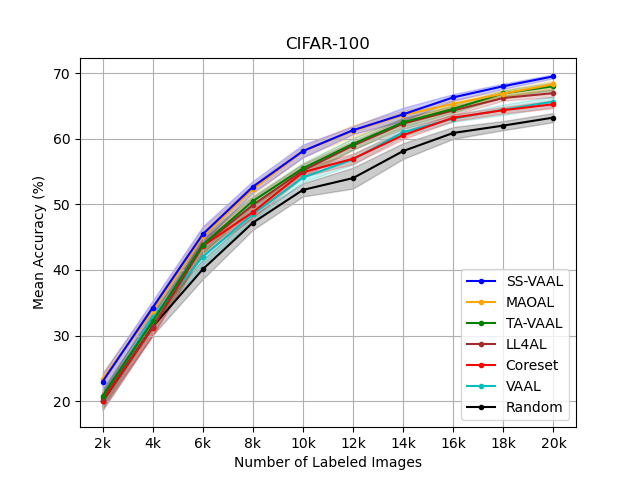}\hspace{-2mm}
  \label{subfig:CIFAR100_results}}
\subfloat[]{
  \includegraphics[width=.35\columnwidth]{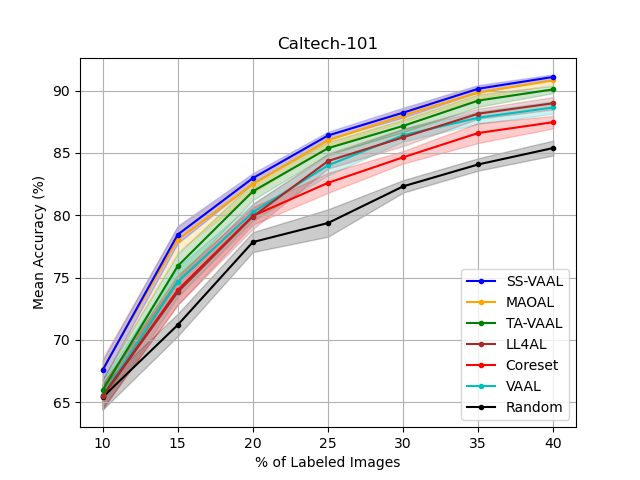}\hspace{-1mm}
  \label{subfig:caltech_results}}\\
\caption{Image classification comparison on the (a) CIFAR-10, (b) CIFAR-100,
and (c) Caltech-101 datasets.}
\label{fig:classification}
\end{figure}




\subsection{Active Learning for Semantic Segmentation}
\textbf{Experimental setup.} To evaluate the effectiveness of our AL approach
in more complex environments, we analyzed the task of semantic segmentation
using Cityscapes \cite{cordts2016cityscapes}, a large-scale dataset of urban
street scene videos. Consistent with the settings in \cite{sinha2019variational},
we utilized the dilated residual network \cite{yu2017dilated} as the semantic
segmentation model. Performance was measured by the mean intersection over union
(mIoU) metric on the Cityscapes validation set. All other experimental settings
were kept consistent with those used in the image classification experiments.

\begin{figure}
\centering
\setlength{\abovecaptionskip}{0.01cm}
\includegraphics[width=0.60\textwidth]{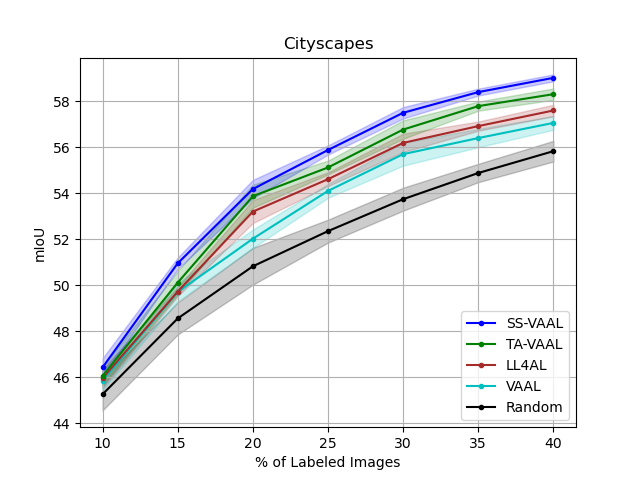}
\caption{Semantic segmentation results on the Cityscapes dataset.}
\label{fig:cityscapes_results}
\end{figure}

\noindent \textbf{Results.} All the compared against methods were averaged
across 3 trials and are shown in Fig.~\ref{fig:cityscapes_results}. Our method
consistently outperforms all the other methods on the task of semantic
segmentation on the Cityscapes dataset as evidenced by its higher mIoU scores.

\subsection{Ablation Study}
To assess the impact of each proposed component, we executed an ablation study
for the classification task on the CIFAR-10, CIFAR-100, and Caltech-101
datasets. The results are presented in the supplementary material, here we
report the main observations. SS-VAAL (w/ ranking only), which refers to the
enhancement of VAAL by integrating the ranking loss-based module, outperforms
VAAL and LL4AL. This confirms the benefits of considering task-related
information in task learning. Moreover, it outperforms TA-VAAL, indicating that
our selection of the listwise ranking method more effectively conveys
task-related information than that of TA-VAAL (Fig.~7 - Fig.~9).

Conversely, SS-VAAL (w/ CAPL only), which entails implementing the proposed
clustering-assisted pseudo-labeling procedure at every stage of model training,
yields a noticeable improvement over all the other methods. This highlights the
effectiveness of exploiting unlabeled data during model training. It also
offers a modest improvement over the SS-VAAL (w/ ranking only) configuration,
implying that leveraging unlabeled data for training contributes more to the
performance improvement than employing alternative means for conveying
task-related information (Fig.~10 - Fig.~12). 
Additionally, we contrast this configuration with SS-VAAL (w/ PL only), which
represents the use of the conventional pseudo-labeling technique. The increase
in performance underscores the effectiveness of our refinement of this method
(Fig.~13 - Fig.~15).

\section{Conclusion}
\label{sec:conclusion}
In this paper we developed key enhancements to both better optimize the use of
vast amounts of unlabeled data during training and incorporate task-related
information. Our approach, SS-VAAL, includes a novel pseudo-labeling algorithm
that allows a model to delve deeper into the data space, thus enhancing its
representation ability by exploiting all unlabeled data in a semi-supervised
way in every AL cycle. SS-VAAL also incorporates a ranking-based loss
prediction module that converts predicted losses into a differentiable ranking
loss. It can be inserted as a rank variable into VAAL's latent space for
adversarial training. Evaluations on image classification and segmentation
benchmarks demonstrate the increased performance of SS-VAAL over
state-of-the-art techniques.

\bibliographystyle{splncs04}
\bibliography{ss-vaal}

\section*{Supplementary Material}
In this supplement we provide image classification results on the ImageNet
dataset and additional experimental results for the ablation study to assess
the impact of each SS-VAAL component.

\subsection{ImageNet Results}
\begin{figure}[h]
\centering
\vspace{-4mm}
\setlength{\abovecaptionskip}{0.05cm}
\includegraphics[width=0.6\columnwidth]{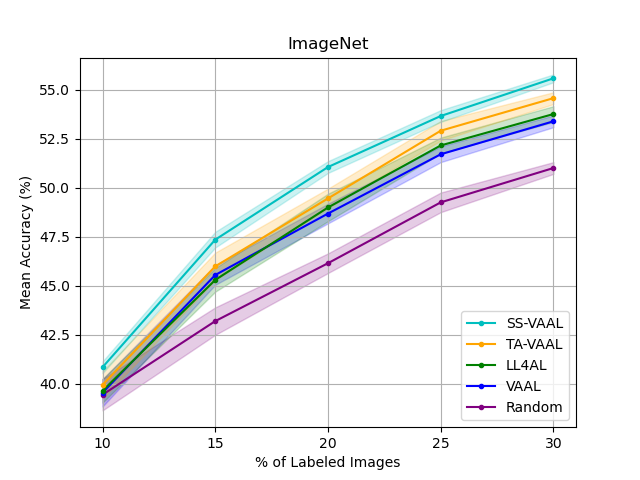}
\caption{Image classification accuracy on the ImageNet dataset.}
\label{fig:imagenet_results}
\vspace{-1mm}
\end{figure}

Fig.~\ref{fig:imagenet_results} presents a performance comparison of our full
methodology against several main competing approaches on the ImageNet dataset.
The results clearly show that SS-VAAL consistently outperforms the others in
every iteration, demonstrating its efficacy and scalability in handling
large-scale datasets.

\subsection{Ablation Study}
\begin{figure}[h]
\centering
\setlength{\abovecaptionskip}{0.11cm}
\includegraphics[width=0.6\columnwidth]{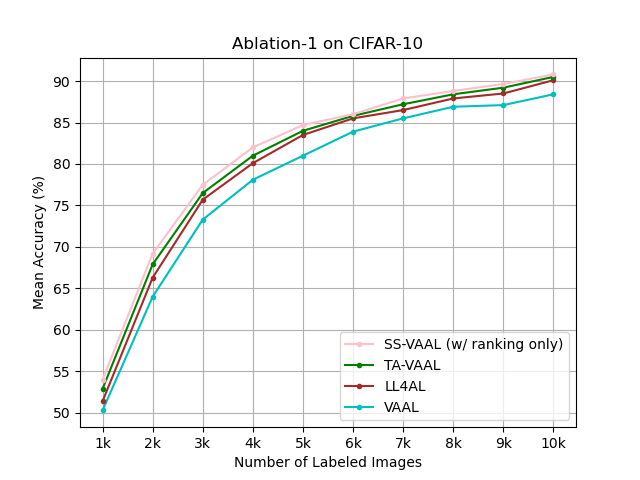}
\caption{Ablation results on analyzing the effect of the ranking component on
the CIFAR-10 dataset.}
\label{fig:cifar10_ablation_1}
\vspace{-7mm}
\end{figure}

\begin{figure}
\centering
\setlength{\abovecaptionskip}{0.11cm}
\includegraphics[width=0.6\columnwidth]{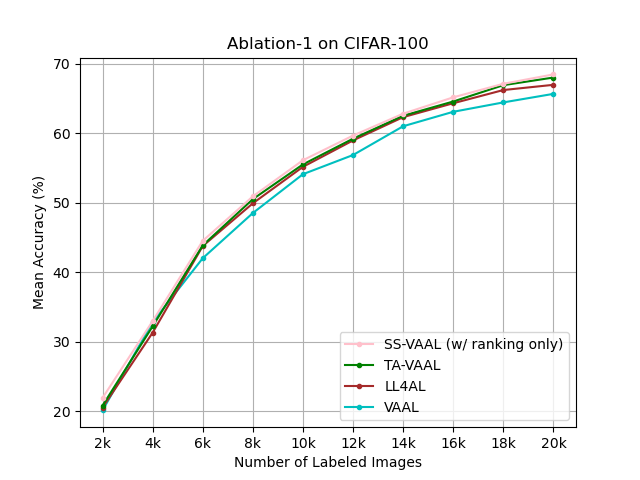}
\caption{Ablation results on analyzing the effect of the ranking component on
the CIFAR-100 dataset.}
\label{fig:cifar100_ablation_1}
\vspace{-5mm}
\end{figure}

\begin{figure}
\centering
\setlength{\abovecaptionskip}{0.11cm}
\includegraphics[width=0.6\columnwidth]{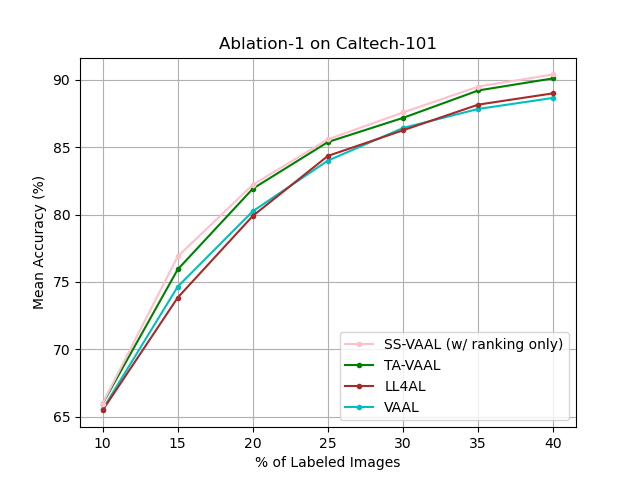}
\caption{Ablation results on analyzing the effect of the ranking component on
the Caltech-101 dataset.}
\label{fig:caltech_ablation_1}
\vspace{-3mm}
\end{figure}

We conducted an ablation study on the image classification task to assess the
impact of each proposed component. SS-VAAL (w/ ranking only) refers to the
enhancement of VAAL by integrating the ranking loss-based prediction module.
According to Fig.~\ref{fig:cifar10_ablation_1} -
Fig.~\ref{fig:caltech_ablation_1}, this configuration outperforms VAAL and
LL4AL, confirming the benefits of considering task-related information in task
learning. Furthermore, this setting also outperforms TA-VAAL, indicating that
our selection of the listwise ranking method more effectively conveys
task-related information than that of TA-VAAL.

\begin{figure}
\centering
\setlength{\abovecaptionskip}{0.11cm}
\includegraphics[width=0.6\columnwidth]{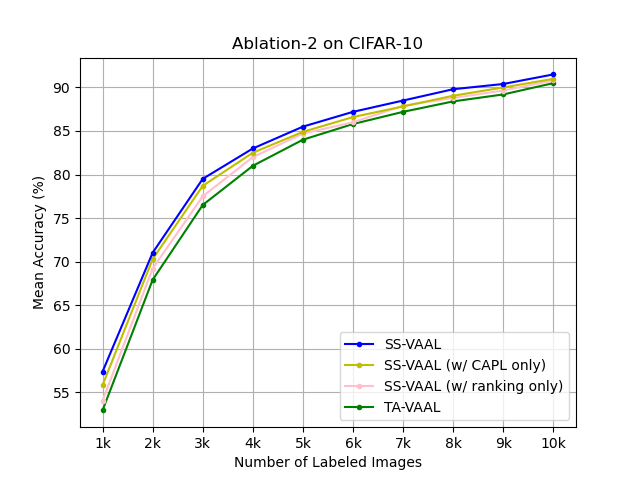}
\caption{Ablation results on analyzing the effect of each component
on the CIFAR-10 dataset.}
\label{fig:cifar10_ablation_2}
\vspace{-3mm}
\end{figure}

\begin{figure}
\centering
\setlength{\abovecaptionskip}{0.11cm}
\includegraphics[width=0.6\columnwidth]{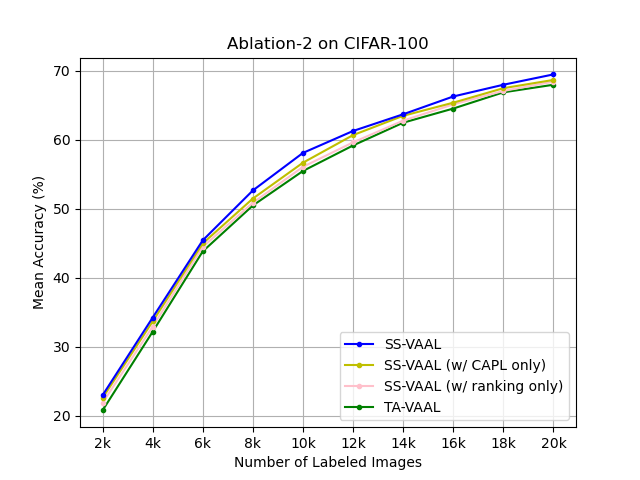}
\caption{Ablation results on analyzing the effect of each component
on the CIFAR-100 dataset.}
\label{fig:cifar100_ablation_2}
\vspace{-3mm}
\end{figure}

\begin{figure}
\centering
\setlength{\abovecaptionskip}{0.11cm}
\includegraphics[width=0.6\columnwidth]{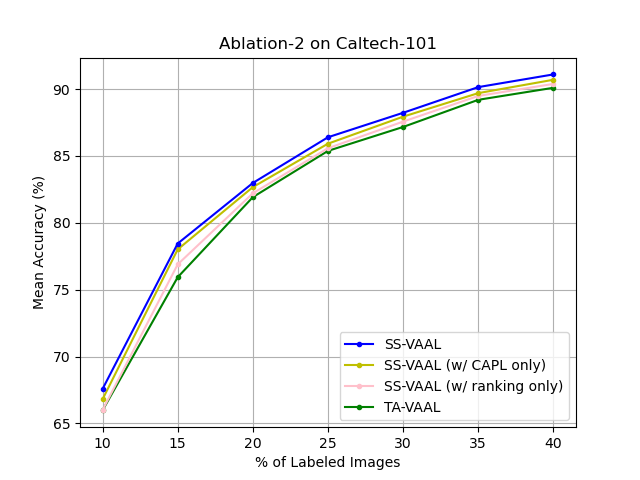}
\caption{Ablation results on analyzing the effect of each component
on the Caltech-101 dataset.}
\label{fig:caltech_ablation_2}
\vspace{-3mm}
\end{figure}

\begin{figure}
\centering
\setlength{\abovecaptionskip}{0.11cm}
\includegraphics[width=0.6\columnwidth]{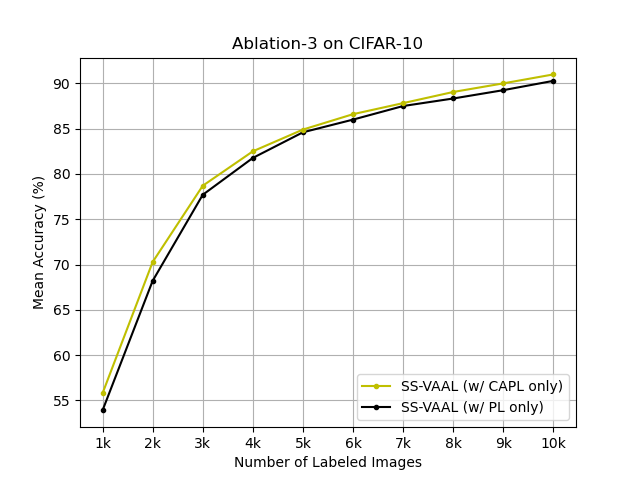}
\caption{Ablation results on analyzing the effect of the pseudo-labeling component
on the CIFAR-10 dataset.}
\label{fig:cifar10_ablation_3}
\vspace{-3mm}
\end{figure}

\begin{figure}
\centering
\setlength{\abovecaptionskip}{0.11cm}
\includegraphics[width=0.6\columnwidth]{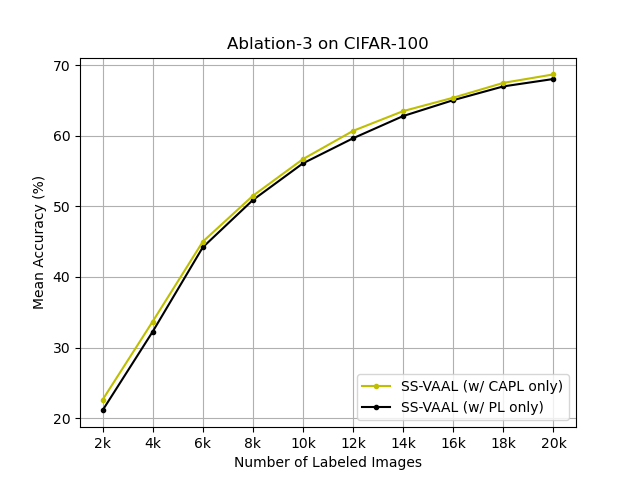}
\caption{Ablation results on analyzing the effect of the pseudo-labeling component
on the CIFAR-100 dataset.}
\label{fig:cifar100_ablation_3}
\vspace{-3mm}
\end{figure}

\begin{figure}
\centering
\includegraphics[width=0.6\columnwidth]{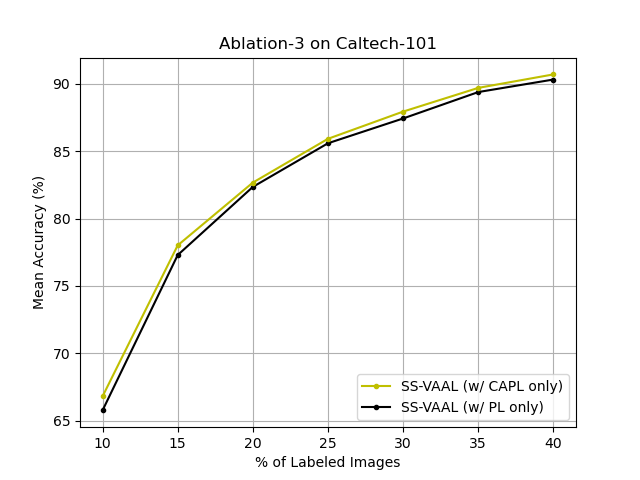}
\caption{Ablation results on analyzing the effect of the pseudo-labeling component
on the Caltech-101 dataset.}
\label{fig:caltech_ablation_3}
\vspace{-3mm}
\end{figure}

On the contrary, SS-VAAL (w/ CAPL only) entails the implementation of the
proposed clustering-assisted pseudo-labeling procedure at every stage of model
training. This setup yields a noticeable improvement over all compared methods,
highlighting the effectiveness of exploiting unlabeled data during model
training.  It also offers a modest improvement over the SS-VAAL (w/ ranking
only) configuration (Fig.~\ref{fig:cifar10_ablation_2} -
Fig.~\ref{fig:caltech_ablation_2}), implying that leveraging unlabeled data for
training contributes more to performance improvement than employing alternative
means for conveying task-related information.  Additionally, we contrast this
configuration with SS-VAAL (w/ PL only), which represents the use of the
conventional pseudo-labeling technique. The enhancement in performance
underscores the effectiveness of our refinement on this method
(Fig.~\ref{fig:cifar10_ablation_3} - Fig.~\ref{fig:caltech_ablation_3}).

\end{document}